\documentclass[10pt,twocolumn,letterpaper]{article}

\usepackage{cvpr}
\usepackage{times}
\usepackage{epsfig}
\usepackage{graphicx}
\usepackage{amsmath}
\usepackage{amssymb}

\usepackage{color}

\usepackage[table]{xcolor}
\usepackage{booktabs}

\usepackage{subcaption}
\usepackage{soul,color}
\usepackage{wrapfig}

\usepackage{multirow}

\usepackage{algpseudocode,algorithm,algorithmicx}

\algrenewcommand\algorithmicrequire{\textbf{Precondition:}}  
\algrenewcommand\algorithmicensure{\textbf{Postcondition:}}

\usepackage{breqn}
\usepackage[pagebackref=true,breaklinks=true,letterpaper=true,colorlinks,bookmarks=false]{hyperref}

\cvprfinalcopy 

\def\cvprPaperID{4325} 

\ifcvprfinal\fi
\begin{document}

\title{Shared Cross-Modal Trajectory Prediction for Autonomous Driving}

\author{Chiho Choi$^{1}$\thanks{Co-first author.}~\quad Joon Hee Choi$^{2*}$~\quad Jiachen Li$^{1,3}$\thanks{Work done during Jiachen's internship at Honda Research Institute}~\quad Srikanth Malla$^{1}$\\
$^1$Honda Research Institute USA\\
$^{2}$Sungkyunkwan University\\
$^{3}$University of California, Berkeley\\
{\tt\small \{cchoi, smalla\}@honda-ri.com\quad \tt\small jhchoi2019@skku.edu\quad \tt\small jiachen_li@berkeley.edu}
}

\maketitle

\begin{abstract}
   Predicting future trajectories of traffic agents in highly interactive environments is an essential and challenging problem for the safe operation of autonomous driving systems. On the basis of the fact that self-driving vehicles are equipped with various types of sensors (\textit{e.g.}, LiDAR scanner, RGB camera, radar, etc.), we propose a \textit{Cross-Modal Embedding} framework that aims to benefit from the use of multiple input modalities. 
   At training time, our model learns to embed a set of complementary features in a shared latent space by jointly optimizing the objective functions across different types of input data. At test time, a single input modality (e.g., LiDAR data) is required to generate predictions from the input perspective (i.e., in the LiDAR space), while taking advantages from the model trained with multiple sensor modalities. An extensive evaluation is conducted to show the efficacy of the proposed framework using two benchmark driving datasets.

\end{abstract}

\section{Introduction}

Future trajectory prediction has become the central challenge to succeed in the safe operation of autonomous vehicles designed to cooperate with interactive agents (\textit{i.e.}, pedestrians, cars, cyclists, etc.). It can benefit to the deployment of applications in autonomous navigation and driving assistance systems with advanced motion planning and decision making. Based on the fact that multi-modal sensors (\textit{e.g.}, LiDAR scanner, RGB cameras, radar, etc.) are equipped in autonomous vehicles, we propose a cross-modal embedding framework that demonstrates the efficacy of the use of multiple sensor data for motion prediction.

\begin{figure}[t]
\begin{center}
\includegraphics[width=1\linewidth]{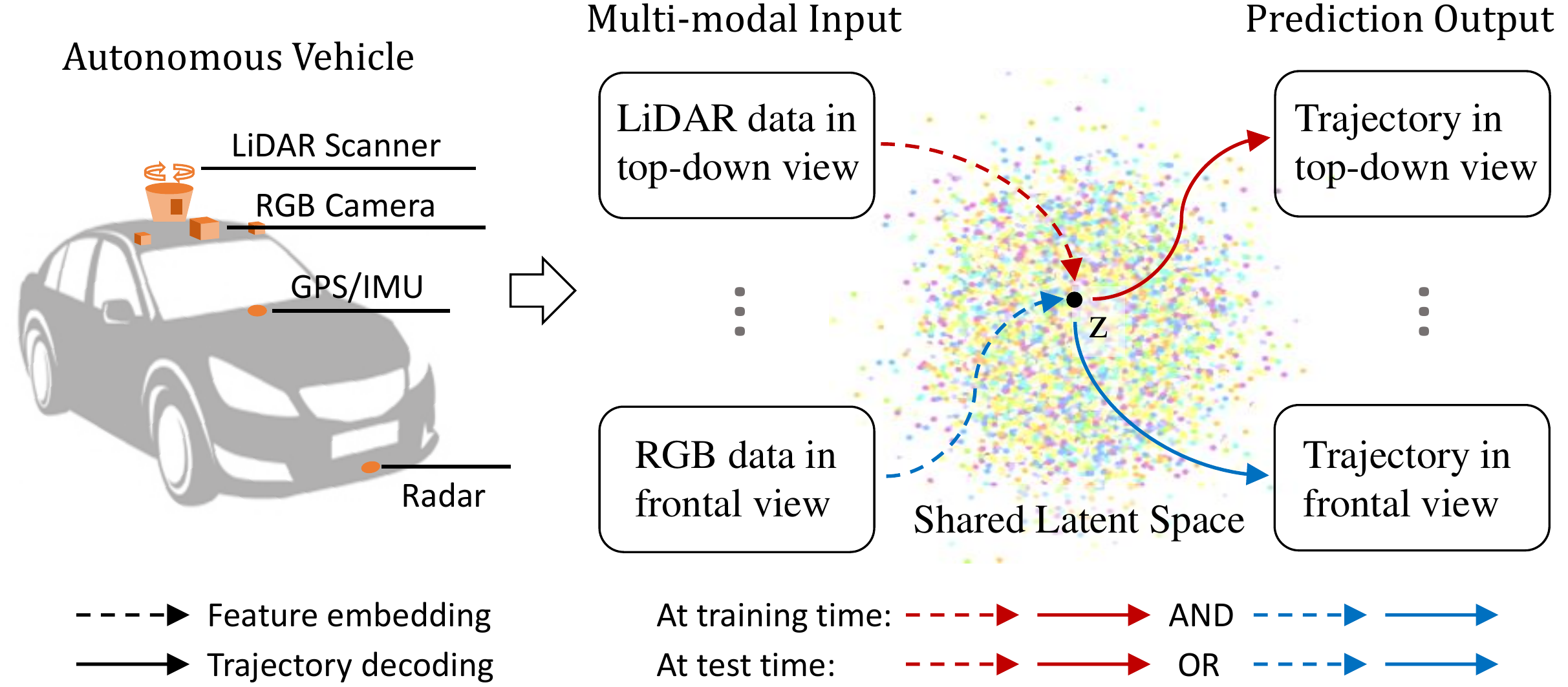}
\end{center}\vspace{-0.3cm}
   \caption{Given a set of multi-modal data (\textit{e.g.}, LiDAR data, RGB images, etc.) obtained from an autonomous vehicle, the model is trained to embed complementary representations of different input modalities into a shared latent space. Output predictions are generated from different perspectives using a latent variable sampled from the learned embedding space. At test time, the proposed method takes a single input modality (\textit{e.g.}, LiDAR data, red-dashed arrow) and predicts the future motion in the same space (\textit{i.e.}, LiDAR-captured world space, red-solid arrow). }\vspace{-0.3cm}
\label{fig:overview}
\end{figure}

Figure~\ref{fig:overview} illustrates an overview of the proposed approach. At training time, we embed multiple feature representations encoded from individual sensor data into a single shared latent space. Our model jointly optimizes the objective functions across different input modalities, so that 
the evidence lower bound of multiple input data over the likelihood can be jointly maximized. We provide a derivation of the objective of shared cross-modal embedding and its implementation using a CVAE-based generative model. At test time, the model takes a single input modality (\textit{e.g.}, LiDAR data) and generates a future trajectory from the input perspective (\textit{i.e.}, top-down view) using a latent variable sampled from the shared embedding space. In this way, we can benefit to the model training from the use of multiple input modalities\footnote{For example, top-down view LiDAR data and frontal view RGB images. However, the input modalities are not limited to these two types but also include stereo images, depth, radar, GPS, and many others equipped in autonomous vehicles, which can provide visual or locational information.}, while keeping the same computational time for trajectory generation as if the single modality had been used. To the best of our knowledge, we are the first to employ multi-modal sensor data from a single framework for trajectory prediction. Note that existing works solve the problem either in top-down view~\cite{lee2017desire, rhinehart2018r2p2, choi2019drogon} with LiDAR data or in frontal view~\cite{yao2018egocentric, bhattacharyya2018long, srikanth2019nemo} with RGB images. 

The proposed framework is clearly distinguishable to studies on a multi-modal pipeline for scene understanding such as detection~\cite{chen2017multi, ku2018joint, liang2019multi}, tracking~\cite{frossard2018end, zhang2019robust}, and semantic segmentation~\cite{ha2017mfnet, valada2019self}. 
They have presented more accurate models by simply fusing different representations extracted from several sensor modalities. The generation of such joint representations, however, would not be desirable in driving automation systems due to the following issues: (i) during inference, it inherently increases the computation time proportional to the number of input modalities used; and (ii) with the anomalous LiDAR data, the model would fail in finding a solution, which is critical to operate self-driving vehicles. 
For the former issue, our proposed cross-modal embedding takes only a single input data during inference and thus does not influence the computational time
, while it still benefits from the model trained with multiple input modalities. In the latter, our model provides alternative prediction solutions in frontal view using the RGB data, which will activate driving assistance functions (\textit{i.e.}, ADAS) for safe vehicle operation, even with a sensor failure.

To this end, we generate multiple modes of future trajectories by sampling several latent variables from the learned latent space. However, such random sampling-based strategy~\cite{lee2017desire,choi2019drogon} is likely to predict similar trajectories, ignoring the random variables while generating predictions from the decoder. This posterior collapse\footnote{We do not carry out any study on mode collapse of GANs or related problems other than posterior collapse of VAEs where our work is built on.} problem of VAEs is particularly critical to future prediction as it mitigates the diverse modes of system outputs. Therefore, 
we introduce a regularizer (i) that pushes the model to rely on the latent variables, predicting diverse modes of future motion; and (ii) that does not weaken the prediction capability of the decoder while preventing the performance degradation. 

We address the following ideas in the proposed method:\vspace{-0.5em} 
\begin{itemize}
    \item The objective of shared cross-modal embedding to jointly approximate a real distribution using multiple input sources is mathematically derived using the Kullback-Leibler divergence (Sec.~\ref{sec:crossmodal}).\vspace{-0.5em} 
    \item Shared cross-modal embedding is implemented based on our derivation to benefit from the use of multiple input modalities, while keeping the same computational time as if the single modality had been used (Sec.~\ref{sec:crossmodal}).\vspace{-0.5em} 
    \item The regularizer is designed for future prediction to mitigate posterior collapse of VAEs and to predict more diverse modes of motion behavior (Sec.~\ref{sec:multimodal}).\vspace{-0.5em}
\end{itemize}
In addition, we design an interaction graph with a graph-level target (Sec.~\ref{sec:social_behavior}), 
introduce a new evaluation metric to measure prediction success (Sec.~\ref{sec:sr}), and propose to use absolute motions in frontal view (Sec.~\ref{sec:input}).

Throughout the paper, we use the word `multi-modality' to denote two different sources. First, \textit{multi-modal input} represents input data obtained from different types of sensors
. Second, \textit{multi-modal prediction} depicts predicted trajectory outputs with multiple variations.

\section{Related Work}
\label{sec:related}



\noindent
\textbf{Pedestrian Trajectory Prediction} A majority of research on trajectory prediction~\cite{alahi2016social,gupta2018social,xu2018encoding,zhang2019sr} has been conducted toward modeling the interactive behavior between humans. These works first encode the temporal information of individual humans and then find their correlation through a social module. Recently, social interactions have been modeled from the graph structure in \cite{vemula2018social,huang2019stgat,mohamed2020social}. Although these methods may be successful in interaction modeling, they overlook the environmental influences that may cause prediction failures in structured environments with stationary obstacles. Therefore, the subsequent work \cite{choi2019looking,kosaraju2019social} takes images as input to constrain their model using scene context.

\vspace{0.3em}
\noindent
\textbf{Vehicle Trajectory Prediction in Top-down View} Similar interaction modules are applied for vehicle trajectory prediction. Some approaches only consider the past motion of road agents~\cite{deo2018multi,park2018sequence,ma2019trafficpredict,li2019interaction}, and thus result in large errors with a complex road environment in traffic scenes. To alleviate such problems, \cite{lee2017desire,rhinehart2018r2p2,li2019conditional,choi2019drogon,rhinehart2019precog,salzmann2020trajectron++} input additional visual cues to condition their model on the road topology. However, they overlook the vehicle interactions against pedestrians, which is most critical to model the natural behavior of vehicles on the road for safe driving. We thus do not limit our scope to `vehicle' trajectories and its interactions. 
Instead, we explicitly discover interactions of heterogeneous entities using the proposed interaction graph.

\vspace{0.3em}
\noindent
\textbf{Vehicle Trajectory Prediction in Frontal View} \cite{bhattacharyya2018long,yao2018egocentric,srikanth2019nemo,malla2020titan} aim to predict the future trajectory of vehicles in a frontal view image space. They predict a target agent's \textit{relative} trajectory with respect to the potential motion of ego-vehicle. Therefore, the predictions are valid only if the accurate ego-future is available. In practice, however, prediction of ego-motion is an another research topic~\cite{huang2019uncertainty} in the transportation domain, which makes hard to simply apply 
such systems to the real world driving applications. Therefore, we predict the \textit{absolute} coordinates of trajectories with no effect of unknown ego-future in frontal view.

\vspace{0.3em}
\noindent
\textbf{Multi-Modal Learning} Learning representations of multiple input modalities have been explored in recent years. As described in~\cite{ngiam2011multimodal}, multi-modal learning can be categorized into three types. \textit{Multi-modal fusion} takes multiple modalities as input and learns their joint representations. Basically, the same set of input types should be provided at test time as in~\cite{jain2016recurrent,yang2016multilayer}. \textit{Cross-modal learning} tries to learn more descriptive representations from one modality when auxiliary modalities are given at training time. 
During inference, the auxiliary modalities are not necessary as in~\cite{gupta2016cross,choi2017learning}. 
\textit{Shared representation learning} learns the representation from one modality and performs the test on the other modality as shown in~\cite{yi2015shared,peng2016cross}. The proposed cross-modal embedding aligns in between cross-modal learning and shared representation learning, similar in spirit to~\cite{aytar2017cross}. We aim to benefit from different modalities that are correlated to each other. However, rather than learning common representations, we train the model to embed different representations into the shared cross-modal latent space. 











\begin{figure*}[!t]
\begin{center}
 \includegraphics[width=0.98\textwidth]{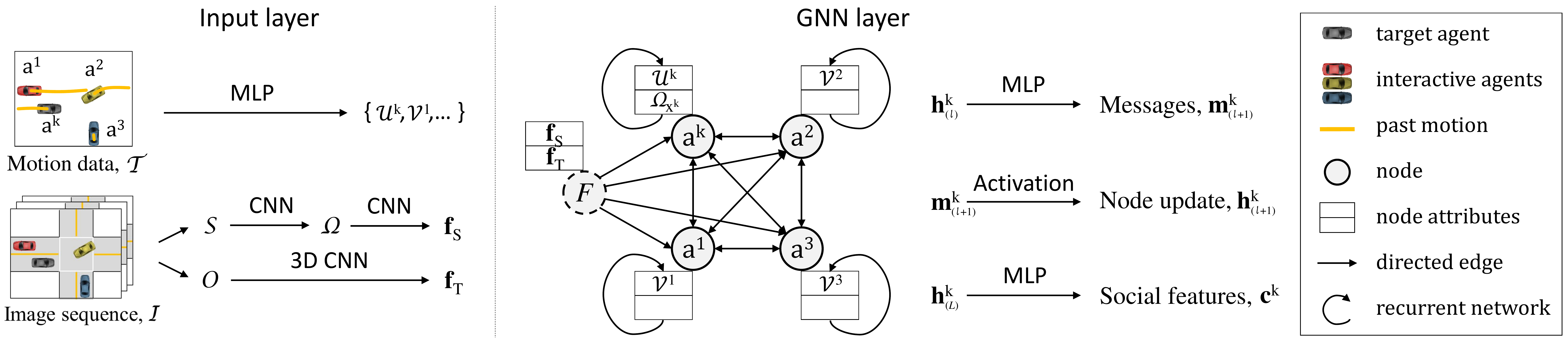}
\end{center}\vspace{-0.5cm}
  \caption{The illustration of the proposed feature encoder. Using the past image sequence, we model spatio-temporal factors given by external environments. The internal and social factors of the target agent is encoded from its past motion as well as surrounding local perceptual context. In the following GNN layer, we model agent-specific social behavior.
  }
\label{fig:fe}\vspace{-0.3cm}
\end{figure*}

\section{Proposed Method}

Given a scenario with the trajectory data $\mathcal{T}=\{{T}^i|\forall i\in \{1,...,K\}\}$ of $K$ traffic agents, we split ${T}^i$ into a past trajectory $\mathbf{x}^i = \{x^i_{t}|\forall t\in\{1,...,\tau\}\}$ for the first $\tau$ observation time steps and a future trajectory $\mathbf{y}^i = \{x^i_{t}|\forall t\in\{\tau+1,...,\tau+\delta\}\}$ for the next $\delta$ time steps, where $x^i_{t}$ represents a 2D position of an arbitrary agent $i$ at time $t$. Assuming that a visual sequence $\mathcal{I}$ is available during $\tau$ observation time steps, we compute the optical flow $\mathcal{O}$ by running TV-L1~\cite{zach2007duality} and segmentation map ${S}$ from DeepLab-V2~\cite{chen2018deeplab} trained on Cityscapes~\cite{cordts2016cityscapes}. Given $\{\mathcal{O},{S}\}$ and $\{\mathbf{x}^i|\forall i\in\{1,...,K\}\}$, our goal is to generate a trajectory prediction $\mathbf{\hat{y}}^k$ of the target agent $k$. To achieve this, we build a feature extraction module in Sec.~\ref{sec:social_behavior} upon graph neural networks (GNNs) in order to learn social behaviors $\mathbf{c}^k$ of the target $k$ toward all other traffic agents (\textit{e.g.}, pedestrians, vehicles, etc.) as well as surrounding road structures. Then, we derive the objective of the proposed shared cross-modal embedding and show its implementation within CVAE in Sec.~\ref{sec:crossmodal}. The encoder $q(\mathbf{z}|\mathbf{y}^k,\mathbf{c}^k)$ is learned to embed $\mathbf{y}^k$ into the latent space, conditioning on the observed social behavior $\mathbf{c}^k$. The following decoder $p(\mathbf{y}^k|\mathbf{z},\mathbf{c}^k)$ reconstructs the future locations ${\mathbf{y}^k}$ using $\mathbf{c}^k$ with a latent sample $\mathbf{z}$. Finally, in Sec.~\ref{sec:multimodal} we provide a solution for mode diversification addressing the posterior collapse issue. 


\subsection{Social Behavior Encoding}
\label{sec:social_behavior}

\noindent
\textbf{Input Layer for External Features} \;\; The importance of external constraints on trajectory prediction is particularly pronounced for traffic agents in driving scenes. To model such environmental influences, the system should be able to recognize each object’s static/dynamic states as well as the semantic context of the scene. 

The image sequence $\mathcal{I}$ captured during the past time steps is used to generate two types of representations: a set of optical flow images $\mathcal{O}$ and a segmentation map ${S}$. The temporal changes of the objects from $\mathcal{O}$ are processed using the 3D convolutional neural network $CNN_{3D}$ by extracting temporal representations $\mathbf{f}_T$ along the time axis:
\begingroup
\setlength\abovedisplayskip{5pt}
\setlength\belowdisplayskip{5pt}
\begin{equation}
    \mathbf{f}_T = CNN_{3D}(\mathcal{O}; W_{T}),
\end{equation}
\endgroup
where $W_{T}$ is the learnable weight parameters.

In addition, a pixel-level segmentation map is obtained at the first time step of the given scenario. Among the estimated labels, we only leave the background structures such as road, sidewalk, vegetation, etc. to extract visual features from the stationary environment. The 2D convolutional neural network $CNN_{2D}$ is used in this stream to take advantage of its spatial feature encoding:
\begingroup
\setlength\abovedisplayskip{5pt}
\setlength\belowdisplayskip{5pt}
\begin{equation}
    \mathbf{f}_S = CNN_{2D}(S; W_{S}),
\end{equation}
\endgroup
where $W_{S}$ is the learnable parameters.

We merge the temporal states $\mathbf{f}_T$ of static/dynamic objects with the spatial features $\mathbf{f}_S$ of the stationary context to generate spatio-temporal features
\begingroup
\setlength\abovedisplayskip{5pt}
\setlength\belowdisplayskip{5pt}
\begin{equation}
    \mathbf{f}_E = \mathbf{f}_T + \mathbf{f}_S.
\end{equation}
\endgroup
We further convert $\mathbf{f}_E\in\mathbb{R}^{d_C\times d_C\times d_E}$ to the external feature matrix $F\in\mathbb{R}^{K\times d_E}$ for the graph. $K$ entities (of size $d_E$) are taken from one of cells in a $d_C\times d_C$ grid of $\mathbf{f}_E$, where the cell location corresponds to each agent $i$'s original pixel location at time $\tau$. For example, an agent shown in the first 32$\times$32 sub-region of an original 256$\times$256 image takes the feature vector from the $(1,1)$-th cell in a 8$\times$8 grid of $\mathbf{f}_E$.

\vspace{0.3em}
\noindent
\textbf{Input Layer for Node Features} \;\; Using the past motion history of traffic agents, we encode the node features. Assuming the task is to predict the future motion of the target agent $k$, we first discover its own intent by preforming the following procedure. The past states $\mathbf{x}^k$ is encoded into high dimensional feature representations $\mathcal{U}_k$ through the multi-layer perceptron (MLP). The encoded features are then combined with the local perception that contains mid-level semantic context $\Omega_{x^k_\tau}$ (nearby areas of ${x}^k_\tau$) from former $CNN_{2D}$. By adding spatial locality, interactions of the target toward the local environment further constrain its motion intent. The subsequent LSTM captures the temporal dependency of motion states on the local environment by
\begingroup
\setlength\abovedisplayskip{5pt}
\setlength\belowdisplayskip{5pt}
\begin{align}
    \mathcal{U}^k &= \text{MLP} \left(\mathbf{x}^k; W_U \right), \nonumber \\
    \mathbf{h}^k_{t+1} &= \text{LSTM} \left(\mathcal{U}^k_{t} + \Omega_{x_\tau^k}, \mathbf{h}^k_{t}; W_K \right),
    \label{eqn:target}
\end{align}
\endgroup
where $W_U$ and $W_K$ is the learnable parameters of MLP and LSTM layer, and $\mathbf{h}^k_{t}$ denotes the hidden state of LSTM at time $t$. We define the last hidden state as $\mathbf{h}^k_{(0)}$ and use it to initialize the node features of the target in the graph.

We run a different feature encoding procedure for the rest of the agents $j\in\{1,...,K\}\backslash\{k\}$ to model their relative motion toward the target agent $k$ as follows:
\begingroup
\setlength\abovedisplayskip{5pt}
\setlength\belowdisplayskip{5pt}
\begin{align}
    \mathcal{V}^j &= \text{MLP} \left(\mathbf{x}^k-\mathbf{x}^j; W_V \right), \nonumber \\
    \mathbf{h}^j_{t+1} &= \text{LSTM} \left(\mathcal{V}^j_{t}, \mathbf{h}^j_{t}; W_J \right),
    \label{eqn:rest}
\end{align}
\endgroup
where $W_V$ and $W_J$ is the learnable parameters of MLP and LSTM, and $\mathbf{h}^j_{t}$ denotes the hidden state of LSTM at time $t$. This process is simple yet effective to infer temporal changes of interactive behavior of individual agents. We use the last hidden state of each agent $j$ as $\mathbf{h}_{(0)}^j$ for the graph. \vspace{0.3em}

\noindent
\textbf{GNN Layer}
The social behavior of the target agent is modeled from each agent's features and external environment features. We define a graph $G = ({H},F)$, where $H\in\mathbb{R}^{K\times d_V}$, $H=\{\mathbf{h}^i|\forall i\in\{1,...,K\}\}$ is a node feature matrix representing $K$ node embeddings of size $d_V$. $F\in\mathbb{R}^{K\times d_E}$ is an external feature matrix $F=\{\mathbf{f}^i|\forall i\in\{1,...,K\}\}$, where each entity represents outside influence on each node in the graph. Following the general message passing phases~\cite{gilmer2017neural}, we construct a GNN architecture:
\begin{equation}
    H_{(l+1)} = M(H_{(l)},F),
\end{equation}
where $M$ is the message propagation function that takes the node feature matrix $H_{(l)}$ updated by $l$ times of the message passing phase. We initialize $H_{(0)}=\{\mathbf{h}^j_{(0)}|\forall j\in\{1,...,K\}\backslash \{k\}\}\cup\{\mathbf{h}^k_{(0)}\}$ using the hidden states obtained from the input layer. 

The proposed GNN structure for social behavior modeling can be considered as a family of pair message
passing neural networks~\cite{battaglia2016interaction}, where the function $M$ takes a concatenation of two nodes as a pair. We design our model on top of this process with an additional graph-level target:
\begingroup
\setlength\abovedisplayskip{5pt}
\setlength\belowdisplayskip{5pt}
\begin{align}
    \mathbf{m}^k_{(l+1)} &= \sum_{i,j} \text{MLP} \bigg(\text{Concat} \big( \mathbf{h}^i_{(l)} + \mathbf{f}^i, \nonumber \\[-0.5ex]
    & \qquad\qquad\qquad \mathbf{h}^j_{(l)} + \mathbf{f}^j, \;\;\mathbf{h}^k_{(l)} \big); W_M \bigg), \nonumber \\
    \mathbf{h}^k_{(l+1)} &= \sigma \left( \mathbf{m}^k_{(l+1)} \right), 
\end{align}
\endgroup
where $W_M$ is the learnable parameters of MLP, Concat(,,) denotes concatenation, $k$ is a target agent, and $i$ and $j$ are the rest of agents. During the message passing phase, the relation between two nodes $i$ and $j$ is encoded with respect to the target node $k$ by considering their external influences $\mathbf{f}^i$, $\mathbf{f}^j$. A summation operation generates messages invariant to the permutation of
the nodes. Then, the features of the target node $\mathbf{h}^k_{(l+1)}$ in the graph are updated by a non-linearity function $\sigma$ such as ReLU using the messages $\mathbf{m}^k_{(l+1)}$. After $L$ updates, the output social behavior features $\mathbf{c}^k$ are generated by another MLP during the readout phase:
\begingroup
\setlength\abovedisplayskip{5pt}
\setlength\belowdisplayskip{5pt}
\begin{equation}
    \mathbf{c}^k = \text{MLP}(\mathbf{h}^k_{(L)};W_R),
\end{equation}
\endgroup
where $W_R$ is the learnable parameters. We use $\mathbf{c}_i^k$ for a certain input $i$. For notational brevity, we drop the target indicator $k$ in the following sections. The input layer and GNN layer is illustrated in Figure~\ref{fig:fe}, and details of the network architecture are shown in the supplementary material.

\subsection{Shared Cross-Modal Framework}
\label{sec:crossmodal}
The main contribution of this work is that we propose a cross-modal embedding framework for future prediction. It aims to benefit from the use of multiple input modalities, while keeping the same computational complexity as if the single data type had been used for trajectory prediction. To implement such functionality, we derive our model within the CVAE framework to embed various types of representations into a single shared latent space. Instead of learning the latent space manifold from a single input, several complementary representations extracted from multiple data sources simultaneously characterize the cross-modal space at training time. By jointly learning the same scenario from different input perspectives, the generative process becomes more descriptive, which results in increasing the performance. At test time, a single modal input is used to sample the latent variables from the learned cross-modal space, taking advantages with other sensor modalities. 

In the followings, we mathematically derive the objective function of shared cross-modal embedding and extend its derivation toward a generative model conditioned on the input observation. 

\vspace{0.3em}
\noindent
\textbf{Joint Optimization}
The objective of cross-modal embedding is to jointly approximate a real distribution $p(\mathbf{z})$ using a posterior $q_i(\mathbf{z}|\mathbf{y}_i)$ of multiple input sources $i\in\{$LiDAR, RGB, ...\}, where $\mathbf{y}_i$ is the sample data point of input modality $i$, and $\mathbf{z}$ is the latent variable. Exploiting the fact that
\begin{equation}
    KL(q(\mathbf{y})||p(\mathbf{y})) = -\int q(\mathbf{y}) \log \left( \frac{p(\mathbf{y})}{q(\mathbf{y})} \right) d\textbf{y} \geq 0,
    \label{eqn:6}
\end{equation}
the Kullback-Leibler (KL) divergence associated with multiple approximates $q_i$ is given by:
\begin{align}
    &\hspace{-0.4cm}\sum_{i} KL(q_i(\mathbf{z}|\mathbf{y}_i)\;||\;p(\mathbf{z}|\mathbf{y}_i)) \nonumber \\
    = & \hspace{0.2cm}\sum_{i} -\int q_i(\mathbf{z}|\mathbf{y}_i) \log \left( \frac{p(\mathbf{z}|\mathbf{y}_i)} {q_i(\mathbf{z}|\mathbf{y}_i)} \right) d\textbf{z} \geq 0.
    \label{eqn:kl1}
\end{align}
By applying Baye's theorem and employing $ \int q_i(\mathbf{z} | \mathbf{y}_i) d\textbf{z} = 1,$
Eqn. (\ref{eqn:kl1}) can be revised as
\begin{align}
    \sum_{i} \bigg( -\int q_i(\mathbf{z} | \mathbf{y}_i) \log \left( \frac{p_i(\mathbf{y}_i|\mathbf{z}) p(\mathbf{z})} {q_i(\mathbf{z}|\mathbf{y}_i)} \right) d\textbf{z} &+ \log p(\mathbf{y}_i) \bigg) \nonumber \\
    & \geq\; 0.
    \label{eqn:kl2}
\end{align}
Using the definition of the KL divergence and expected value and simple math, Eqn. (\ref{eqn:kl2}) is converted to
\begin{align}
    \log & \left( \prod_{i} p(\mathbf{y}_i) \right) \geq \sum_{i} \bigg( - KL \left( q_i(\mathbf{z}|\mathbf{y}_i) || p(\mathbf{z}) \right) \nonumber \\
    &\qquad\qquad\quad\; + \mathbb{E}_{ \sim q_i(\mathbf{z}|\mathbf{y}_i)}[\log p_i(\mathbf{y}_i|\mathbf{z})] \bigg).
    \label{eqn:kl3}
\end{align}
Therefore, maximizing the evidence lower bound (ELBO) of multiple input data over the likelihood jointly maximizes their evidence probability. \vspace{0.3em}

\noindent
\textbf{Cross-Modal Embedding} The proposed cross-modal embedding framework is trained to jointly learn the shared latent space conditioned on multiple input observations such as $\mathbf{c}_{i\in\{\mathrm{LiDAR,RGB,etc.}\}}$. The variational lower bound of the log-likelihood can be extended as a conditional form by
\begin{multline}
    \log \bigg(\displaystyle\prod_i p(\mathbf{y}_i|\mathbf{c}_i)\bigg)
    \geq\displaystyle\sum_i\bigg(-{KL}(q_i(\mathbf{z}|\mathbf{y}_i,\mathbf{c}_i)||p(\mathbf{z}|\mathbf{c}_i))\\+\mathbb{E}_{ \sim q_i(\mathbf{z}|\mathbf{y}_i,\mathbf{c}_i)}[\log p_i(\mathbf{y}_i|\mathbf{z},\mathbf{c}_i)]\bigg),
\end{multline} 
where $q_i(\mathbf{z}\vert \mathbf{y}_i,\mathbf{c}_i)$ and $p_i(\mathbf{y}_i|\mathbf{z},\mathbf{c}_i)$ is implemented as a pair of an encoder and decoder for $i$-th input modality following the reparameterization trick of CVAE. $\mathbf{c}_i$ is the conditional observation. \textbf{The full derivation is provided in the supplementary material.} We draw the loss to minimize the negative ELBO while training the model as follows:
\begin{multline}
  \mathcal{L}_{E} = \displaystyle\sum_i \bigg({KL}(q_i(\mathbf{z}\vert \mathbf{y}_i,\mathbf{c}_i)\Vert p(\mathbf{z}\vert \mathbf{c}_i))\\ - \mathbb{E}_{\sim q_i(\mathbf{z}\vert \mathbf{y}_i,\mathbf{c}_i)}[\log p_i(\mathbf{y}_i|\mathbf{z},\mathbf{c}_i)] \bigg).
  \label{eqn:cvae_all}
\end{multline}
The network parameters of the encoder are learned to minimize the KL divergence between the prior distribution $p(\mathbf{z}\vert \mathbf{c}_i)$ and the approximates $q_i(\mathbf{z}\vert \mathbf{y}_i,\mathbf{c}_i)$. The second term is the log-likelihood of samples, which is considered as the reconstruction loss of the decoder. The decoder generates trajectories using the latent variables $\mathbf{z}$ sampled from the prior that is modeled as Gaussian distribution $\mathbf{z}\thicksim\mathcal{N}(0,\textnormal{I})$.

\subsection{Multi-modal Prediction}
\label{sec:multimodal}

In practice, the optimization of VAE and its variants is challenging itself because of the posterior collapse problem. The strong autoregressive power of the decoder often ignores the random variable $\mathbf{z}$ sampled from the learned latent space. Thus, the output is dominantly generated using the conditional input $\mathbf{c}$, still satisfying the minimization of the KL divergence and maximization of the log-likelihood in Eqn.~(\ref{eqn:cvae_all}). Such a problem alleviates the multi-modal nature of future prediction where multiple plausible trajectories are generated given the same past motion. To address posterior collapse, we consider the following challenges: (i) our technique helps to generate diverse responses from the decoder, which enables multi-modal prediction and (ii) it does not physically weaken the decoder to alleviate its prediction capability. 
In this sense, we design an auxiliary regularizer that makes the decoder to rely on the latent variable. 

At training time, we assume that there exist $N$ modes of trajectories for each query. Then, the latent variables $\mathbf{z}_n \thicksim q(\mathbf{z}_n\vert \mathbf{y},\mathbf{c}) = \mathcal{N} (\mu,\sigma^2)$ are sampled from the normal distribution with the mean $\mu$ and variance $\sigma^2$, where $n\in\{1,...,N\}$. We consider the trajectories generated using these latent variables as $N$ modes of prediction outputs. To maximize the diversity among predictions, the pair-wise similarity is evaluated using Gaussian kernel by
\begin{equation}
    K = \exp\left(-\frac{D(\mathbf{\hat{y}}_i,\mathbf{\hat{y}}_j)}{2\sigma_G^2}\right),
\end{equation}
where $D$ measures a distance between predictions $\mathbf{\hat{y}}_i$ and $\mathbf{\hat{y}}_j$ with $i,j\in\{1,...,N\}$ and $\sigma_G^2$ is the hyper-parameter of this kernel function. We find a maximum similarity $K_{max}$ and minimize it during training. The regularizer then enforces the model 
to maximize the diversity among $N$ predicted trajectories through the optimization without losing the prediction capability of the decoder.

As a result, the total objective function of the proposed approach is drawn as follows:
\begin{equation}
    \mathcal{L}_{Total} = \mathcal{L}_{E} + \lambda\displaystyle\sum_i K_{max,i}
    \label{eqn:total}
\end{equation}
where $i\in\{$LiDAR,RGB,...$\}$ is an indicator for input data modalities and $\lambda$ balances multi-modality and accuracy ($\lambda=10$ is used). To optimize the first term in Eqn.~(\ref{eqn:total}), we find $\mathbf{\hat{y}}_n$ of the mode $n$ that best reconstructs the ground truth $\mathbf{y}$. In this way, the log-likelihood in Eqn.~(\ref{eqn:cvae_all}) encourages the decoder to generate accurate results, while preserving the mode diversity with the regularizer.





\section{Experiments}

\subsection{Input Modalities}\label{sec:input}
Any set of sensory data can be used as input to the proposed framework. For demonstration, however, we use two exemplary data types that are easily accessible from the existing benchmark datasets: (i) \textit{LiDAR data} provide 3D scanning of the surrounding environment. Using 3D point clouds, we project every single point onto the ground plane in top-down view and predict trajectories of traffic agents in the LiDAR-captured world coordinates. (ii) \textit{RGB images} captured from a frontal-facing camera provide rich and dense representations. We predict the trajectories from the egocentric perspective in the image space. Unlike relative trajectories in~\cite{Yagi_2018_CVPR,yao2018egocentric}, we propose to predict trajectories using the absolute locations, 
eliminating the effect of uncertain ego-future~\cite{srikanth2019nemo}. We provide its details with our data preparation in the supplementary material.

\subsection{Datasets and Evaluation Metrics}\label{sec:sr}
\noindent
\textbf{Datasets} Two benchmark driving datasets (KITTI~\cite{geiger2013vision} and H3D~\cite{patil2019h3d}) are used to evaluate the proposed approach comparing to self-generated baselines and state-of-the-art methods. The KITTI dataset was introduced for trajectory forecast in \cite{lee2017desire} to predict future motions of road agents in top-down view, and then \cite{yao2018egocentric} found their future locations in frontal view using this dataset. As in~\cite{lee2017desire}, we generate a set of trajectory segments with 6 $sec$ long (2 $sec$ for observation and 4 $sec$ for prediction) using Road and City scenes in the Raw subset. We divide all videos into five sets and conduct 5-fold cross validation, following the split of~\cite{choi2019looking}. In addition, the H3D~\cite{patil2019h3d} dataset is used to further validate the proposed approach on heterogeneous agents in highly congested urban environments. For evaluation, we divide 160 scenarios of H3D into the training (75\%) and test set (25\%) and use the same observation / prediction time as KITTI. 

\vspace{0.3em}
\noindent
\textbf{Metrics} For the performance comparison, we mainly follow the standard evaluation metrics: 
\begin{itemize}
    \item \textit{Average Distance Error (ADE)} is computed using L2 distance between the predicted trajectory and the ground truth for a certain time duration.
    \item \textit{Final Distance Error (FDE)} shows L2 distance between the predicted location and the ground truth at a certain time step.
\end{itemize}
Both ADE and FDE are reported with 1 $sec$ interval at future time steps. For multi-modal prediction, we sample 20 trajectories and find the best one with a minimum ADE at 4 $sec$ in future. Note that the single- and multi-modal models are respectively denoted by a different suffix \_S and \_M.

In addition, we introduce a new metric that measures the rate of prediction success:
\begin{itemize}
    \item \textit{Success Rate (SR)} finds the fraction of scenarios where L2 distance between the predicted endpoint and ground truth is within a certain threshold value $\epsilon$. 
\end{itemize}
Under the assumption that the prediction would be successful if the error at the endpoint is within a certain threshold, this metric plots how many scenarios can be considered as `success prediction'. SR thus is a more practical evaluation metric that tests the overall robustness of the algorithm.

\begin{table}[!t]
\centering
\resizebox{0.47\textwidth}{!}{\begin{tabular}{c|c|l||cccc}
\hline
\multicolumn{3}{c||}{Component}&\multirow{2}{*}{1.0 $sec$}&\multirow{2}{*}{2.0 $sec$} &\multirow{2}{*}{3.0 $sec$}&\multirow{2}{*}{4.0 $sec$}\\\cline{1-3}
Env&Soc&~~Mul&&&&\\ 
\hline 
\hline
-&-&~~~~~-&0.37 / 0.64&0.69 / 1.47&1.20 / 3.01&1.94 / 5.32\\
+&- &~~~~~-&0.38 / 0.65&0.68 / 1.39&1.16 / 2.96&1.87 / 4.97\\
-&+ &~~~~~-&0.33 / 0.55&0.61 / 1.31&1.09 / 2.80&1.79 / 4.92\\
+&+ &~~~~~-&0.31 / 0.51 &0.53 / 1.07&0.92 / 2.36&1.53 / 4.35\\
\hline
+&+ &+ Fus&{0.20} / {0.35} &{0.42} / {1.00}&{0.82} / {2.31}&{1.45} / {4.38}\\
+&+ &+ Emb&{0.20} / {0.36} &{0.42} / {1.00}&{0.82} / {2.29}&{1.44} / {4.33}\\
\hline
\end{tabular}%
}\vspace{-0.2cm}
\caption{Ablation study on the KITTI~\cite{geiger2013vision} dataset. ADE / FDE is reported in \textit{meters}. Refer to Sec.~\ref{sec:ablation} for description.  }
\label{tbl:ablation}\vspace{-0.5cm}
\end{table}

\subsection{Ablative Study} 
\label{sec:ablation}
We first demonstrate our design choices through ablative studies conducted in top-down view using KITTI. We evaluate the baseline models on the following components: 
\begin{itemize}
    \item \textbf{Env}: External features ($\textbf{f}_S$ and $\textbf{f}_T$); 
    \item \textbf{Soc}: Social influences of other agents; 
    \item \textbf{Mul}: Multi-modal learning. \textbf{Fus}: multi-modal fusion with feature aggregation, \textbf{Emb}: proposed shared cross-modal embedding.
\end{itemize} 
Table~\ref{tbl:ablation} compares ADE and FDE of six baseline models that are designed by adding (+) or dropping (-) these components. When one or more of information is missing, a significant performance drop is observed. The error of the model without any components is particularly larger than others by a huge margin. By considering environmental influences (\textbf{Env}), the performance improves toward long-term prediction ($4 sec$). It clearly demonstrates the effectiveness of the environmental constraints on more distant areas. We observe that adding \textbf{Soc} outperforms previous baselines, which implies the role of social behavior encoding for trajectory prediction with the significant improvement at the short-term time steps. The impressive error drop is found by taking both \textbf{Env} and \textbf{Soc} into account. It demonstrates the validity of the proposed feature extractor. We highlight the efficacy of the use of multiple sensor modalities from \textbf{Mul}, where both \textbf{Fus} and \textbf{Emb} further improve the performance. Interestingly, the proposed cross-modal embedding (\textbf{Emb}) achieves even lower error than fusion-based counterpart (\textbf{Fus}). 
It indicates that our model benefits from complementary input modalities with cross-modal embedding, even though a single input data is used during inference.

Additionally, we show the efficacy of the proposed regularizer for multi-modal prediction in the bottom of Table~\ref{tbl:kitti3}. Without the proposed regularizer (\textbf{S-CM\_10 w/o reg}), the performance improvement of the model with 10 samples is minimal against single-modal prediction (\textbf{S-CM\_1}), which is interpreted as a posterior collapse problem. However, the model with the regularizer (\textbf{S-CM\_10}) highly improves the accuracy generating diverse output responses. We conclude that the proposed regularizer can ease posterior collapse for future prediction.


\begin{table}
\centering
\resizebox{0.48 \textwidth}{!}{
\begin{tabular}{l|c|llll}
\hline
Method &$N$& 1.0 $sec$& 2.0 $sec$&3.0 $sec$&4.0 $sec$\\
\hline
\hline 
\textit{State-of-the-art}&&&&\\
~~Const-Vel~\cite{scholler2019constant} &1& 0.34 / 0.56 &0.85 / 1.79&1.60 / 3.72&2.55 / 6.24\\
~~S-LSTM~\cite{alahi2016social}&1 & 0.53 / 1.07&1.05 / 2.10&1.93 / 3.26&2.91 / 5.47\\
~~Gated-RN~\cite{choi2019looking}&1 & 0.34 / 0.62&0.70 / 1.72&1.30 / 3.34&2.09 / 5.55\\
~~DESIRE~\cite{lee2017desire} &1&~~~~~-~ / 0.51&~~~~~-~ / 1.44&~~~~~-~ / 2.76&~~~~~-~ / 4.45\\
~~DESIRE~\cite{lee2017desire}&20 &~~~~~-~ / \textbf{0.28}&~~~~~-~ / 0.67&~~~~~-~ / 1.22&~~~~~-~ / 2.06\\
~~S-GAN~\cite{gupta2018social}&20 & 0.29 / 0.43&0.67 / 1.34&1.26 / 2.94&2.07 / 5.22 \\
~~S-STGCNN~\cite{mohamed2020social}&20&0.21 / 0.36&0.38 / 0.70 & 0.59 / 1.31& 0.82 / 2.14 \\
~~Trajectron++~\cite{salzmann2020trajectron++}&20&0.19 / 0.33&0.34 / 0.65 & 0.53 / 1.18 & 0.78 / {1.96} \\
\hline
\textit{Ours}&&&&\\
~~S-CM\_1&1&{0.20} / {0.36} &{0.42} / {1.00}&{0.82} / {2.29}&{1.44} / {4.33}\\
~~S-CM\_10 w/o reg&10&0.20 / 0.35 &0.40 / 0.96&0.77/ 2.06&1.33 / 4.04 \\
~~S-CM\_10&10&{0.18} / 0.31 &{0.32} / {0.61}&{0.49}/ {1.09}&{0.75} / {1.99} \\
~~S-CM\_20&20&\textbf{0.17} / 0.29 &\textbf{0.29} / \textbf{0.53} & \textbf{0.42}/ \textbf{0.83}&\textbf{0.61} / \textbf{1.57} \\
 \hline
\end{tabular}
}\vspace{-0.2cm}
\caption{Quantitative comparison (ADE / FDE in \textit{meters}) of our approach with the state-of-the-art methods. The KITTI dataset~\cite{robicquet2016learning} is used to predict trajectories in top-down view.  $N$ denotes the number of samples used.
   }
\label{tbl:kitti3}
\end{table}

\begin{table}
\centering
\resizebox{0.48 \textwidth}{!}{
\begin{tabular}{l|c|llll}
\hline
Method &$N$&1.0 $sec$&2.0 $sec$&3.0 $sec$&4.0 $sec$\\
\hline 
\hline
\textit{State-of-the-art}&&&&\\
~~Conv1D$^*$~\cite{Yagi_2018_CVPR}&1&24.38 / 44.13&~~~~~-~ / ~-&~~~~~-~ / ~-&~~~~~-~ / ~-\\
~~FVL$^*$~\cite{yao2018egocentric}&1&17.88 / 37.11&~~~~~-~ / ~-&~~~~~-~ / ~-&~~~~~-~ / ~-\\
~~Const-vel~\cite{scholler2019constant}&1&5.88 / 9.42&13.23 / 26.03&22.13 / 45.99&31.90 / 68.03\\
~~S-GAN~\cite{gupta2018social}&1&7.54 / 10.51&13.39 / 23.74&21.37 / 43.82&31.76 / 71.55\\
~~S-GAN~\cite{gupta2018social}&20&6.96 / 9.58&12.25 / 21.42&19.48 / 39.66&28.89 / 65.02\\
\hline
\textit{Ours}&&&&\\
~~S-CM\_1 w/o Emb&1&4.17 / 7.85&8.22 / 17.68&13.63 / 30.48&19.63 / 44.97\\
~~S-CM\_1&1&4.17 / 7.35&8.21 / 17.64&13.59 / 30.41&19.59 / 44.92\\
~~S-CM\_10&10&{3.25} / {5.57}&{5.71} / {10.74}&~~{8.25} / {15.74}&{11.22} / {24.85}\\
~~S-CM\_20&20&\textbf{3.19} / \textbf{5.42}&\textbf{5.52} / \textbf{10.02}&~~\textbf{7.51} / \textbf{12.26}&~~\textbf{9.59} / \textbf{19.76}\\
\hline
\end{tabular}
}\vspace{-0.2cm}
   \caption{ADE / FDE is evaluated in $pixels$. The KITTI~\cite{geiger2013vision} dataset is used to predict trajectories in frontal view. $*$ denotes the evaluation on relative motion from ego-vehicle. $N$ denotes the number of samples used.
   }
\label{tbl:kitti2}\vspace{-0.2cm}
\end{table}

\subsection{Quantitative Results}
We first compare the performance of the proposed approach with the state-of-the-art methods using KITTI. In Table~\ref{tbl:kitti3}, we observe from single-modal prediction ($N$=1) that our \textbf{S-CM\_1} outperforms all compared single-modal approaches including social interaction oriented methods~\cite{alahi2016social} as well as scene context oriented methods \cite{lee2017desire,choi2019looking}. 
For multi-modal prediction, the proposed approach (\textbf{S-CM\_10}) with $N$=10 already achieves overall lower ADE and FDE than other competitors in top-down view trajectory forecast. By sampling $N$=20 modes, we improve FDE at 4.0 $sec$ over 19\% against  \cite{salzmann2020trajectron++}. 

Using the same cross-modal model, we examine the frontal view prediction capability in Table~\ref{tbl:kitti2}. Note that Conv1D~\cite{Yagi_2018_CVPR} and FVL~\cite{yao2018egocentric} predicts relative motion with respect to the future ego-motion. Their poor performance might be caused by the prediction difficulties with unknown ego-future. 
Although the proposed method (S\_CM\_1) further improved the accuracy without affecting the inference time, the effect seems less significant compared to that shown in top-down view (Table~\ref{tbl:ablation}). Our insight is as follows: (i) the use of complementary features obtained from different input modality is not as impactful as it was for top-down prediction; and (ii) the performance improvement achieved by other aspects (\textit{e.g.}, social behavior, semantic context, etc.) is already exceptional in frontal view, which makes the improvement with embedding less compelling. Nevertheless, the proposed method with cross-modal embedding generally shows higher accuracy against others. 


\begin{table}[!t]
\centering
\resizebox{0.48 \textwidth}{!}{
\begin{tabular}{l|c|llll}
\hline
Method&$N$ & 1.0 $sec$& 2.0 $sec$&3.0 $sec$&4.0 $sec$\\
\hline 
\hline
\textit{State-of-the-art}&&&&\\
~~Const-Vel~\cite{scholler2019constant}&1 & {0.18} / 0.26 &0.34 / 0.60&0.52 / 1.03&0.74 / 1.54\\
~~S-LSTM~\cite{alahi2016social}&1 &0.26 / 0.41&0.49 / 0.92&0.72 / 1.53&1.01 / 2.32\\
~~S-GAN~\cite{gupta2018social}&1 &0.27 / 0.37&0.45 / 0.77&0.68 / 1.29&0.94 / 1.91\\
~~S-GAN~\cite{gupta2018social}&20 &0.26 / 0.35&0.44 / 0.72&0.65 / 1.24&0.90 / 1.84\\
~~Gated-RN~\cite{choi2019looking}&20 &0.18 / 0.32&0.32 / 0.64&0.49 / 1.03&0.69 / 1.56\\
~~STGAT~\cite{huang2019stgat}&20 &0.24 / 0.33 & 0.34 / 0.48 & 0.46 / 0.77 & 0.60 / 1.18\\
~~S-STGCNN~\cite{mohamed2020social}&20& 0.23 / 0.32& 0.36 / 0.52 & 0.49 / 0.89 & 0.73 / 1.49\\
~~Trajectron++~\cite{salzmann2020trajectron++}&20 &0.21 / 0.34&0.33 / 0.62&0.46 / 0.93&0.71 / 1.63\\
~~EvolveGraph~\cite{li2020evolvegraph}&20 &0.19 / 0.25 & 0.31 / 0.44 & 0.39 / 0.58 & 0.48 / 0.86\\
\hline
\textit{Ours}&&&&\\
~~S-CM\_1&1&{0.14} / {0.25} &{0.27} / {0.54}&{0.43} / {0.95}&{0.62} / {1.45}\\
~~S-CM\_10&10&{0.12} / {0.21} &{0.21} / {0.37}&{0.30} / {0.61}&{0.42} / {0.96}\\
~~S-CM\_20&20&\textbf{0.11} / \textbf{0.19} &\textbf{0.18} / \textbf{0.30}&\textbf{0.25} / \textbf{0.46}&\textbf{0.34} / \textbf{0.77}\\
 \hline
\end{tabular}}\vspace{-0.2cm}
   \caption{Quantitative results (ADE / FDE) are reported in \textit{meters}. We use H3D~\cite{patil2019h3d} to evaluate the proposed method in top-down view. $N$ denotes the number of samples used.
   }
\label{tbl:h3d3}\vspace{-0.2cm}
\end{table}

\begin{table}
\centering
\resizebox{0.48 \textwidth}{!}{
\begin{tabular}{l|c|llll}
\hline
Method &$N$&1.0 $sec$&2.0 $sec$&3.0 $sec$&4.0 $sec$\\
\hline
\hline
\textit{State-of-the-art}&&&&\\
~~Const-vel~\cite{scholler2019constant}&1&13.15 / 19.22&24.64 / 44.13&38.18 / 74.75&53.38 / 110.07\\
~~S-GAN~\cite{gupta2018social}&1&12.91 / 17.05&20.57 / 33.53&29.70 / 54.41&40.71 / 84.51\\
~~S-GAN~\cite{gupta2018social}&20&12.38 / 16.26&19.67 / 31.86&28.30 / 51.55&38.88 / 80.55\\
\hline
\textit{Ours}&&&&\\
~~S-CM\_1&1&8.69 / 16.06&16.52 / 33.25&25.68 / 54.91&36.29 / 82.05\\
~~S-CM\_10&10&{6.62} / {11.36}&{10.69} / {18.12}&{14.25} / {24.51}&{18.22} / {36.92}\\
~~S-CM\_20&20&\textbf{6.25} / \textbf{10.51}&~~\textbf{9.61} / \textbf{15.26} & \textbf{12.07} / \textbf{18.66}&\textbf{15.05} / \textbf{30.56}\\
\hline
\end{tabular}
}\vspace{-0.2cm}
\caption{Our approach is evaluated on ADE / FDE (in \textit{pixels}) using the H3D~\cite{patil2019h3d} dataset. The proposed absolute motions (with ego-future elimination) are used to compute errors in frontal view. $N$ denotes the number of samples used.}
\label{tbl:h3d2}\vspace{-0.4cm}
\end{table}

\begin{figure*}[!t]
  \centering%
  \vspace{-0.3cm}
  \begin{subfigure}[b]{0.98\linewidth}
    \includegraphics[width=\linewidth]{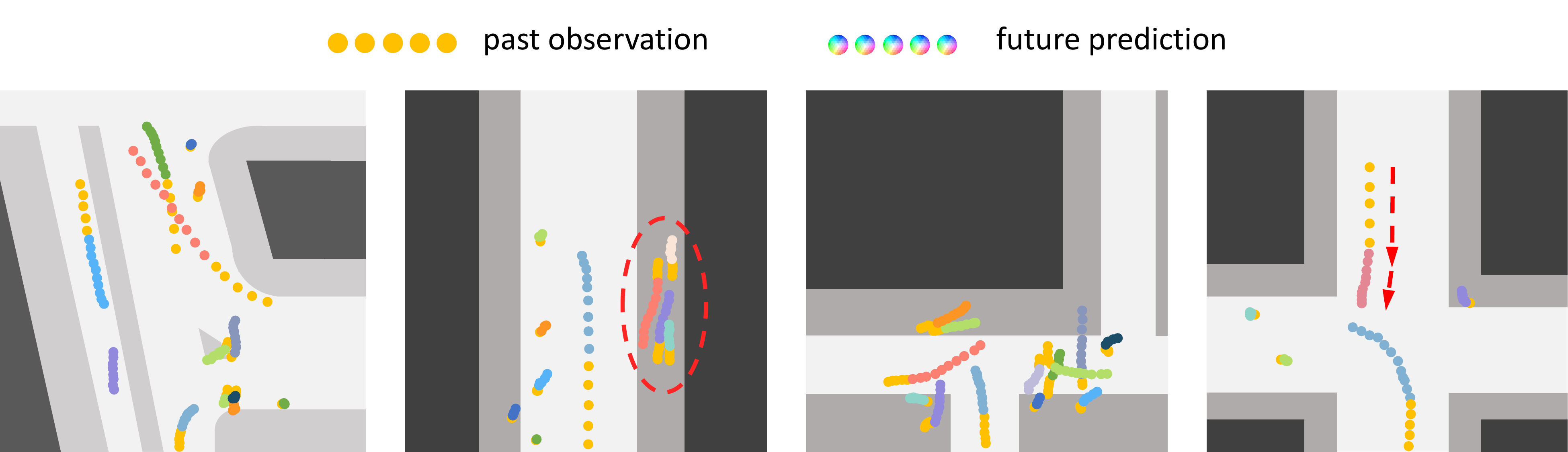}
  \end{subfigure}
  \vspace{-0.2cm}
  \caption{We visualize the top-1 prediction from highly interactive scenarios among heterogeneous traffic agents, such as human-human, human-vehicle, and vehicle-vehicle interactions.
  }
  \label{fig:qual2}\vspace{-0.4cm}
\end{figure*}

\begin{figure}
\begin{center}
    \begin{subfigure}[b]{0.17\textheight}
        \includegraphics[width=\textwidth]{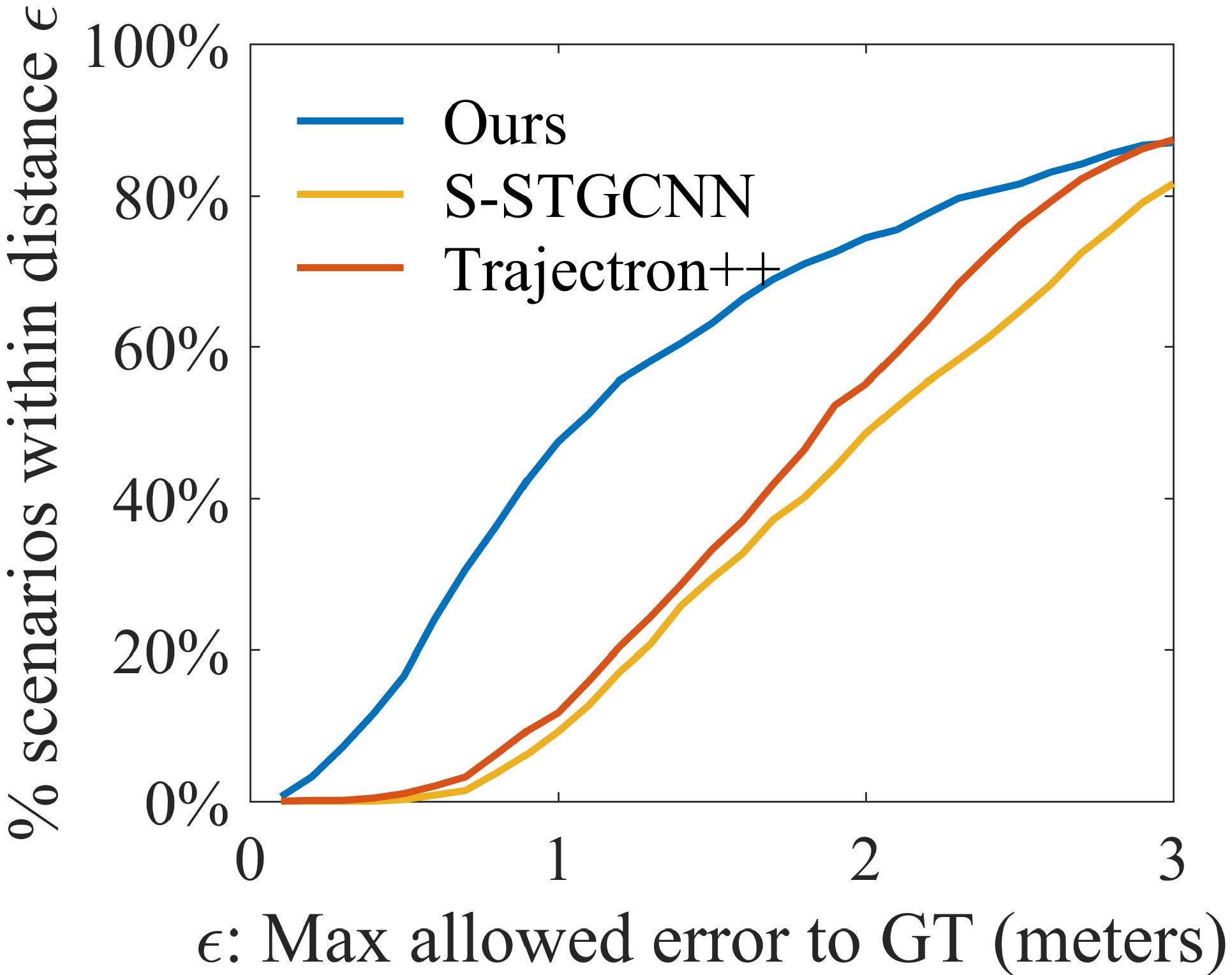}
        \caption{}
        \label{fig:sr_kitti}
    \end{subfigure}\quad
    \begin{subfigure}[b]{0.17\textheight}
        \includegraphics[width=\textwidth]{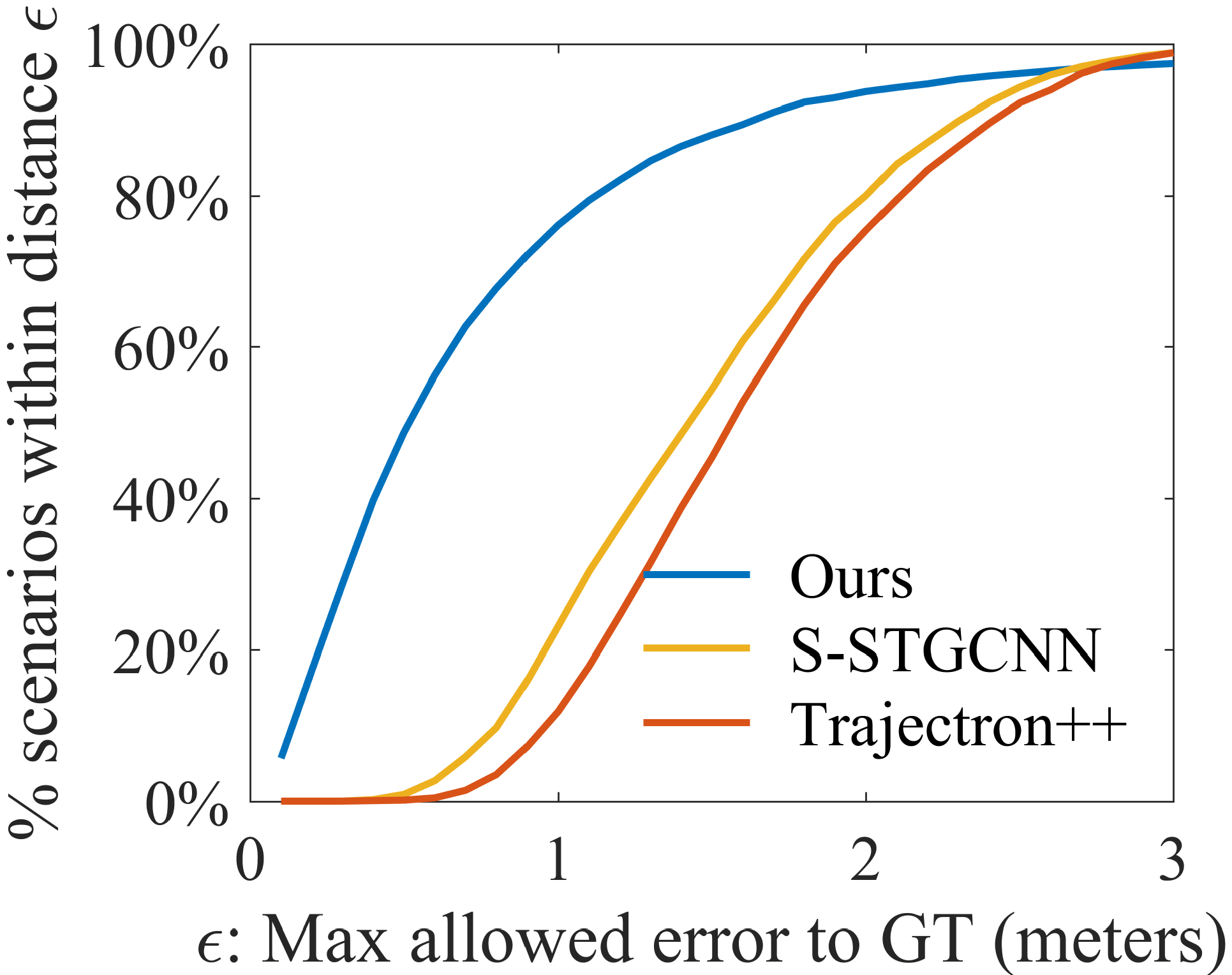}
        \caption{}
        \label{fig:sr_h3d}
    \end{subfigure}
\end{center}\vspace{-0.6cm}
\caption{The success rate (SR) is plotted as a fraction of success prediction scenarios with respect to maximum allowed distance error, which is indicative of overall robustness of algorithms. SR is evaluated in top-down view on FDE at 4.0 $sec$. (a) KITTI and (b) H3D.}
\label{fig:sr}\vspace{-0.4cm}
\end{figure}

We further evaluate our work using the H3D dataset in its highly congested environments. In top-down view as in Table~\ref{tbl:h3d3}, we found that 
the proposed model with a single sample (\textbf{S-CM\_1}) already achieves the lower error than most of methods including very recent graph-based models (\cite{mohamed2020social} w/o scene and \cite{salzmann2020trajectron++} w/ scene). Our explicit modeling of relational interactions together with cross-modal embedding enables us to explore more discriminative behavior representations over these graph-based methods. 
The performance is further improved by sampling multiple predictions with the regularizer (\textbf{S-CM\_20}). Compared to the best state-of-the-art method~\cite{li2020evolvegraph} that finds the dynamic evolution of interactions, our work improves the performance over 10\% on FDE at 4.0 $sec$. Such lower errors demonstrates the generation of highly diverse yet acceptable future motions using our model, considering the road topology.

Subsequently, we evaluate our trajectory prediction framework for the task of frontal view forecast. In Table~\ref{tbl:h3d2}, we observe that the performance of our single-modal prediction model (\textbf{S-CM\_1}) is on par with multi-modal prediction model of Social-GAN (\textbf{S-GAN} with $N$=20). It implies that the prediction capability of the proposed framework is being at the level of the state-of-the-art. The significant improvement of error from our multi-modal prediction model \textbf{S-CM\_20} further demonstrates the effectiveness of our objective function for optimization.

\subsection{Evaluation with Success Rate}

The standard evaluation metrics such as ADE and FDE do not capture the success or failure of predictions. We thus introduce SR that plots the proportion of scenarios that can be considered as ‘successful prediction’ with respect to the definition of \textit{success}. We use the error threshold $\epsilon$ on the x-axis and measure the rate of success scenarios by FDE at 4.0 $sec$. 
Figure~\ref{fig:sr} compares our approach with two state-of-the-art methods~\cite{mohamed2020social,salzmann2020trajectron++}. 
We observe from \ref{fig:sr_kitti} that our approach performs better than others in terms of the correctness of predictions. 
Assuming that the real driving application is designed with a small prediction tolerance ($\epsilon=1.5 m$), our model is more reliable and credible with considerably higher success rate (63\% compared to \cite{salzmann2020trajectron++} of 33\% or \cite{mohamed2020social} of 29\%). 
We also plot SR using the H3D dataset in \ref{fig:sr_h3d}, which indicates that our prediction model can achieve much smaller errors in the majority of scenarios. Our method shows consistently higher success rate, validating the robustness of our prediction capability.

\subsection{Qualitative Results in Top-down View}
Figure~\ref{fig:qual2} visualizes the top-1 prediction result of the proposed approach. Each scenario contains the heterogeneous agents (\textit{i.e.}, cars, bus, pedestrians, cyclist, etc.) interactive one to another. We robustly forecast their future motions by taking advantages of the proposed social behavior modeling and cross-modal embedding. In between pedestrians, our approach models their motion behaviors and generates socially acceptable trajectories (dotted oval in the second column). In the last column, our model predicts that the car would turn left, which influences the behavior of on-coming vehicle that slows its speed (\textit{i.e.}, yielding; dotted arrow). We conclude that the proposed graph accordingly considers relational interactions while predicting future motions. We provide the results of 20 prediction samples as well as qualitative results in frontal view in the supplementary material.

\section{Conclusion}
We proposed a solution to future trajectory forecast in driving scenarios. Assuming that the multiple sensory data is available for autonomous driving, our approach can benefit from the model trained using multiple input modalities. First, the GNN-based feature encoder extracts social behaviors of the target agent, considering its interactions toward all other traffic agents as well as surrounding road structures. Then, the relational behaviors obtained from multiple perspectives are embedded into a shared cross-modal latent space. We provided its derivation that jointly optimizes objective functions using the generative variational models. Finally, we designed an auxiliary regularizer to ease the posterior collapse problem for future prediction. We analyzed the significance of the proposed approach through the extensive evaluation, showing the improvement of the performance against the state-of-the-art methods.

{\small
\bibliographystyle{ieee_fullname}
\bibliography{egbib}
}

\clearpage

\appendix
\section*{Supplementary Material}


\section{Derivation of Cross-Modal Embedding}
In this section, we provide a complete derivation of the objective function of shared cross-modal embedding. Based on Eqn. (\ref{eqn:6}), the KL divergence approximating the conditional posteriors is written as follows:
\begin{multline}
    KL(q(\mathbf{z}|\mathbf{y},\mathbf{c})||p(\mathbf{z}|\mathbf{y},\mathbf{c}))\\
    = -\int q(\mathbf{z}|\mathbf{y},\mathbf{c}) \log \left( \frac{p(\mathbf{z}|\mathbf{y},\mathbf{c})}{q(\mathbf{z}|\mathbf{y},\mathbf{c})} \right) d\textbf{z}
    \geq 0,
     \label{eqn:kl-2-c}
\end{multline}
where $\mathbf{y}$ and $\mathbf{c}$ respectively denotes the data point and condition. Assuming multiple input sources $i\in\{$LiDAR, RGB, ...\} are available, we derive the objective for an arbitrary number of modalities, yielding
\begin{multline}
    \hspace{-0.3cm}\displaystyle\sum_i KL\bigg(q_i(\mathbf{z}|\mathbf{y}_i,\mathbf{c}_i)||p(\mathbf{z}|\mathbf{y}_i,\mathbf{c}_i))\bigg)\\
    = \displaystyle\sum_i-\int q_i(\mathbf{z}|\mathbf{y}_i,\mathbf{c}_i) \log \left( \frac{p(\mathbf{z}|\mathbf{y}_i,\mathbf{c}_i)}{q_i(\mathbf{z}|\mathbf{y}_i,\mathbf{c}_i)} \right) d\textbf{z}\geq 0,
     \label{eqn:kl-2-c}
\end{multline}
which is still under the premise that the KL divergence is non-negative. We apply Baye's theorem 
\begin{equation}
    p(\mathbf{z}|\mathbf{y},\mathbf{c}) = \frac{p(\mathbf{y}|\mathbf{z},\mathbf{c})p(\mathbf{z}|\mathbf{c})}{p(\mathbf{y}|\mathbf{c})}
\end{equation}
to Eqn. (\ref{eqn:kl-2-c}) and employ 
\begin{equation}
    \int q(\mathbf{z}|\mathbf{y},\mathbf{c})d\mathbf{z} = 1,
\end{equation}
yielding
\begin{multline}
    \hspace{-0.3cm}\displaystyle\sum_i KL\bigg(q_i(\mathbf{z}|\mathbf{y}_i,\mathbf{c}_i)||p(\mathbf{z}|\mathbf{y}_i,\mathbf{c}_i))\bigg)\\
    = \displaystyle\sum_i\bigg(-\int q_i(\mathbf{z}|\mathbf{y}_i,\mathbf{c}_i) \log \left( \frac{p_i(\mathbf{y}_i|\mathbf{z},\mathbf{c}_i)p(\mathbf{z}|\mathbf{c}_i)}{q_i(\mathbf{z}|\mathbf{y}_i,\mathbf{c}_i)} \right) d\textbf{z}\\
    +\log p(\mathbf{y}_i|\mathbf{c}_i)\bigg)\geq 0.
\end{multline}
By transferring integral to the other side, we get
\begin{multline}
    \displaystyle\sum_i\log p(\mathbf{y}_i|\mathbf{c}_i)\\
    \hspace{-0.3cm}\geq\displaystyle\sum_i\bigg(\int q_i(\mathbf{z}|\mathbf{y}_i,\mathbf{c}_i)\log \left( \frac{p(\mathbf{z}|\mathbf{c}_i)}{q_i(\mathbf{z}|\mathbf{y}_i,\mathbf{c}_i)} \right) d\textbf{z}\\
    +\int q_i(\mathbf{z}|\mathbf{y}_i,\mathbf{c}_i)\log p_i(\mathbf{y}_i|\mathbf{z},\mathbf{c}_i)d\textbf{z}\bigg).\\
\end{multline}
Therefore, the evidence lower bound of multiple data over the conditional likelihood is given by
\begin{multline}
    \log \bigg(\displaystyle\prod_i p(\mathbf{y}_i|\mathbf{c}_i)\bigg)
    \geq\displaystyle\sum_i\bigg(-{KL}(q_i(\mathbf{z}|\mathbf{y}_i,\mathbf{c}_i)||p(\mathbf{z}|\mathbf{c}_i))\\+\mathbb{E}_{ \sim q_i(\mathbf{z}|\mathbf{y}_i,\mathbf{c}_i)}[\log p_i(\mathbf{y}_i|\mathbf{z},\mathbf{c}_i)]\bigg),
\end{multline} 
which completes the proof.

\begin{figure*}
\begin{center}
 \includegraphics[width=1\textwidth]{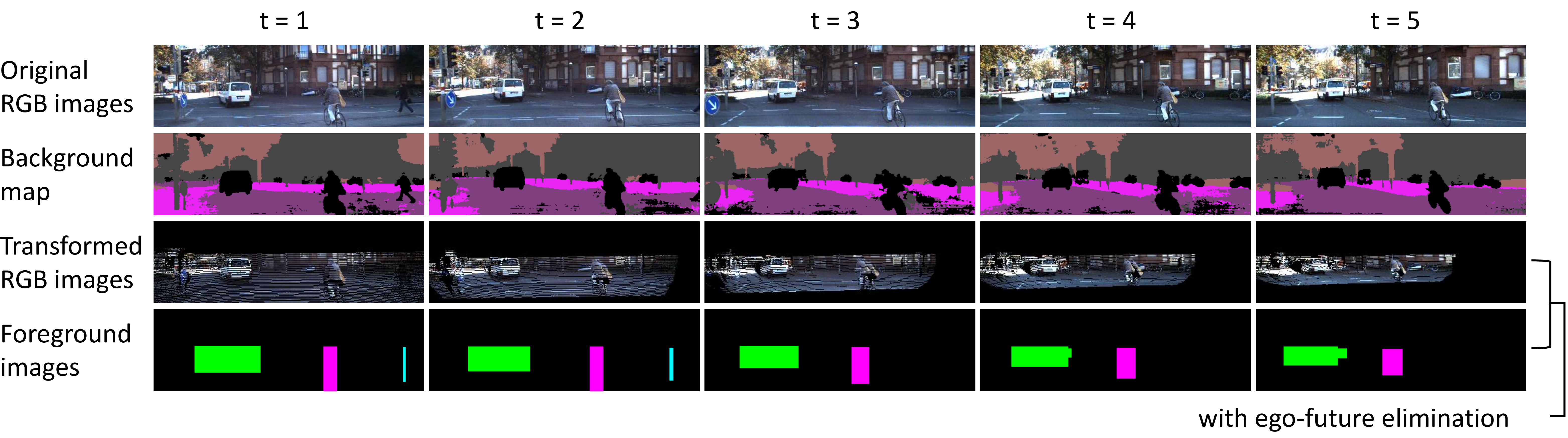}
\end{center}\vspace{-0.5cm}
  \caption{Frontal view input.
  }
\label{fig:front}
\end{figure*}

\begin{figure*}
\begin{center}
 \includegraphics[width=0.8\textwidth]{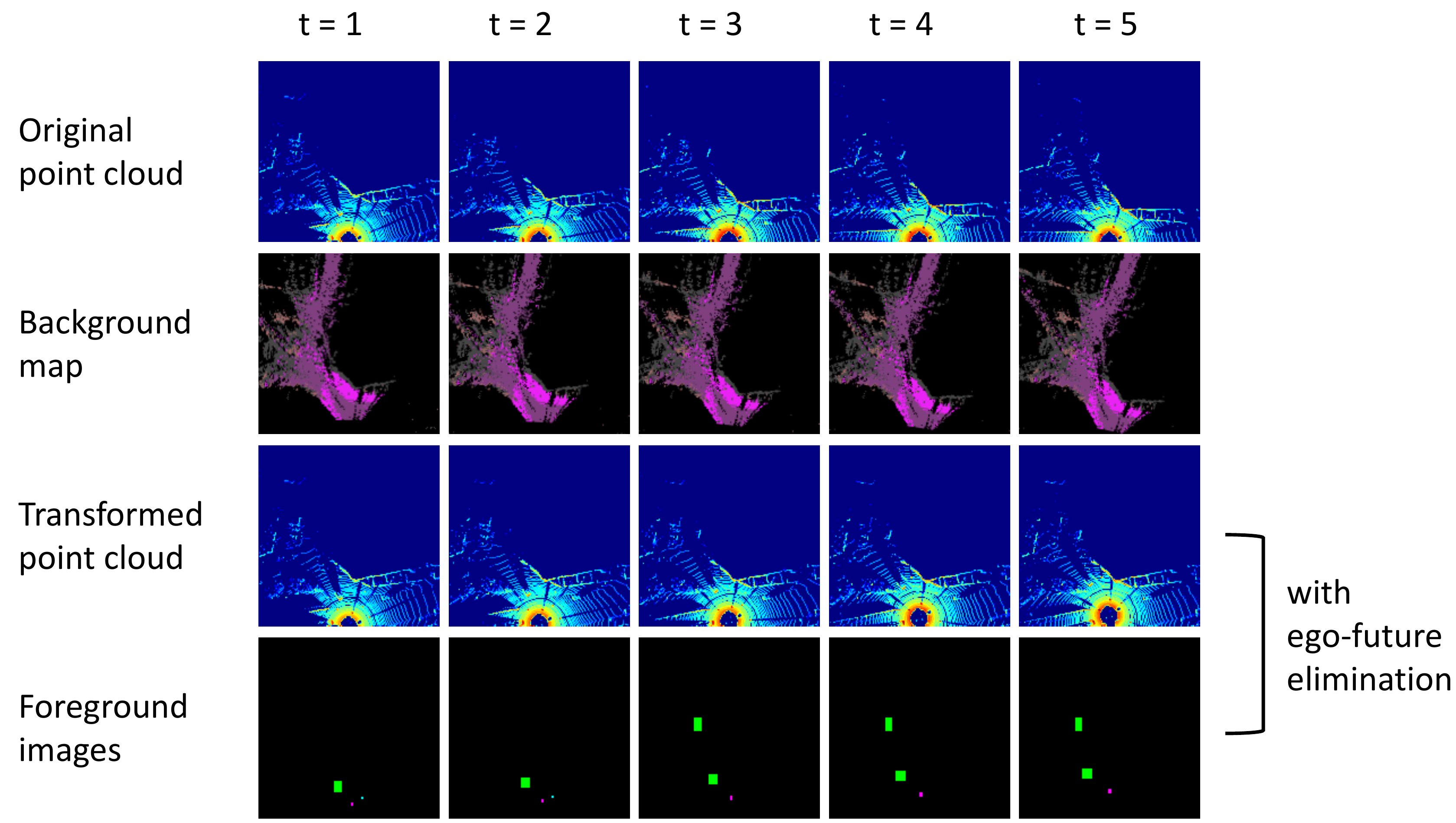}
\end{center}
  \caption{Top-down view input.
  }
\label{fig:topdown}
\end{figure*}

\section{Dataset Preprocessing}
\label{sec:preprocessing}



\subsection{Frontal View Input}
\noindent
\textbf{Segmentation Map}
Frontal view RGB images (top row in Figure.~\ref{fig:front}) are used to estimate the semantic labels in the scene. We first run the DeepLab-V2
~\cite{chen2018deeplab} model trained on the Cityscapes dataset \cite{cordts2016cityscapes}. Then, we leave the background labels with stationary structures (\textit{i.e.}, road, sidewalk, building, etc.) to get the background map as shown in the second row of Figure~\ref{fig:front}. 
For the static stream, we directly use the background map at the first observation time of each scenario as segmentation input $S$ to the model.

\noindent
\textbf{Ego-future Elimination}
To eliminate the effect of ego-future from frontal view prediction, we introduce the absolute coordinate where the motion of traffic agents is not influenced by the ego-vehicle. To eliminate the ego-motion for a dynamic stream, we define the new coordinates at the first observation time step $t=1$ for every trajectory segments $\mathbf{T}^i$ (of length $\tau+\delta$), and all foreground objects are projected into this space. For this, we conduct the following procedure. First, the point cloud in the world coordinates is projected into the image space to grab the corresponding RGB information. Next, we transform the point cloud to the first frame using GPS/IMU position estimates. Finally, the transformed point cloud is projected into the blank image and displayed using previously acquired RGB information. The transformed RGB images generated from this procedure are shown in the third row of Figure~\ref{fig:front}.

\noindent
\textbf{Optical Flow Images}
We first locate the ground truth bounding box in the image space. Then, we go through ego-future elimination steps using these images with corresponding point clouds. Output foreground images (last row in Figure~\ref{fig:front}) are used to compute optical flow $\mathcal{O}$ by running the TV-L1~\cite{zach2007duality} algorithm. The positions of each traffic agent is also computed by this procedure. Images have a dimension of 414 $\times$ 125.

\subsection{Top-down View Image}
\noindent
\textbf{Segmentation Map}
From each trajectory segment, we grab the segmentation label for the point cloud from the RGB-based background map. Then, the point cloud is transformed onto the top-down image space that is discretized with a resolution of 0.5 $m$ as illustrated in the second row of Figure~\ref{fig:topdown}. We use the map at time $t=1$ as $S$ in top-down view.

\begin{figure*}[!t]
\begin{center}
 \includegraphics[width=1\textwidth]{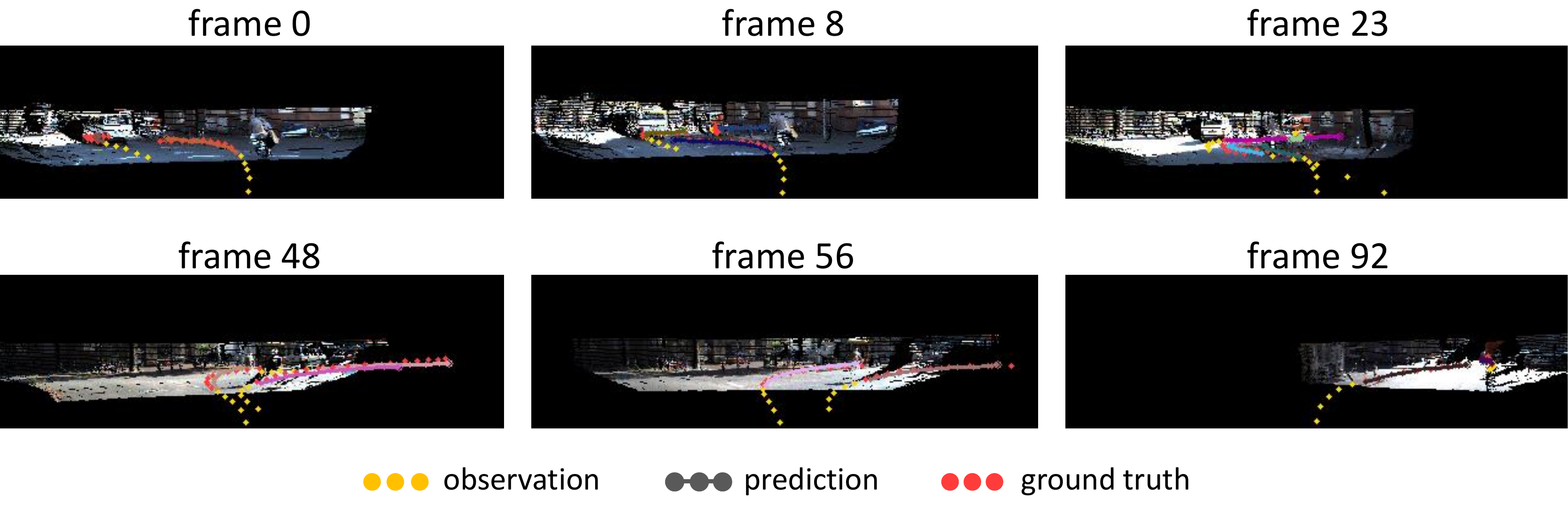}
\end{center}
  \caption{Additional qualitative results evaluated using KITTI in frontal view.
  }
\label{fig:kitti}
\end{figure*}

\begin{figure}
\begin{center}
 \includegraphics[width=0.48\textwidth]{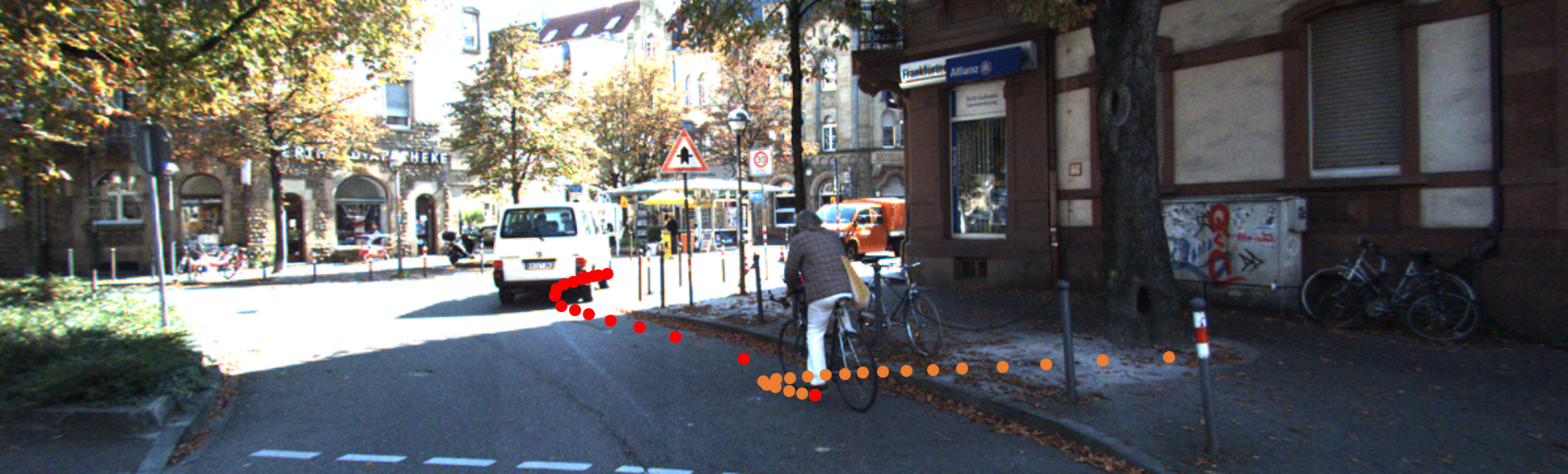}
\end{center}\vspace{-0.5cm}
  \caption{Comparison between absolute (red trajectory) and relative (orange trajectory) motions of a cyclist.
  }
\label{fig:motions}
\end{figure}

\noindent
\textbf{Ego-future Elimination}
Similar to frontal view images, we first transform each point cloud to the local coordinates at $t = 1$ using GPS/IMU position estimates. The transformed point clouds are projected into the top-down view image space with a resolution of 0.5 $m$. The third row of Figure~\ref{fig:topdown} shows the transformed point clouds that are created using the original point cloud in the first row of Figure~\ref{fig:topdown}. 

\noindent
\textbf{Optical Flow Images}
The ground truth bounding box of objects are first processed to eliminate the effect of ego-future. Then, they are drawn in the top-down view image space as displayed in the last row of Figure~\ref{fig:topdown}. The final output has a dimension of $160\times 160$ that corresponding to 80 $m$ to the longitudinal direction and $\pm 40$ $m$ to the lateral direction.

\subsection{Absolute vs Relative Motions in Frontal View}
In Figure~\ref{fig:motions}, we compare the absolute motions introduced in the proposed method with the relative motions used in other approaches. The absolute trajectory (in red dots) is intuitive by eliminating the ego-future. In contrast, the relative trajectory (in orange dots) is not interpretable without extra information of future ego-motion that should be separately predicted (as in \cite{huang2019uncertainty}), considering the uncertainty~\cite{srikanth2019nemo}. 

\section{Additional Evaluation}
\subsection{Qualitative Results in Frontal View}
We additionally show the qualitative results of the proposed method in frontal view using the KITTI~\cite{geiger2013vision} dataset. As in Figure~\ref{fig:kitti}, our frontal view prediction is based on the absolute locations (with ego-future elimination) in the local coordinates of the first frame of each trajectory segment. The proposed approach recognizes the road layouts and accurately predict interactive future motions of different types of road agents. Note that we visualize top-1 prediction in this figure. 

\subsection{Quantitative Results in Top-down View} Table~\ref{tbl:missrate} shows an additional study on extra metrics. We provide FDE at 4.0 $sec$ with standard deviation computed from five experiments. It demonstrates that our approach consistently provides robust prediction capabilities with lower standard deviation. 

\subsection{Additional Top-down View Videos} We provide additional video clips (\textit{scenario1.mp4, scenario2.mp4, scenario3.mp4}) to visualize entire 20 predictions of all traffic agents in the scene. The videos are generated using three interactive scenarios of the H3D dataset. Please check attached videos for additional qualitative evaluation (Color codes: Blue - past observation, Green - ground truth, Red - our predictions).


\begin{table}[!t]
\centering
{\begin{tabular}{l|c}
\hline
{Method}&{FDE $(m)\downarrow$}\\
\hline
\hline
Const-Vel~\cite{scholler2019constant}&1.54\\
S-STGCNN~\cite{mohamed2020social}&1.49{\footnotesize{$\pm0.0239$}}\\
Trajectron++~\cite{salzmann2020trajectron++}&1.63{\footnotesize{$\pm0.0126$}}\\
\hline
Ours&\textbf{0.77}{\footnotesize{$\pm\mathbf{0.0039}$}}\\ 
\hline
\end{tabular}
}
\caption{Additional study using H3D. FDE at 4.0 $sec$ with standard deviation is reported in \textit{meters}. }
\label{tbl:missrate}
\end{table}

\section{Implementation}


We share details of our model architectures. 

\subsection{Social Behavior Encoding}
\vspace{0.5em}
\noindent
\textbf{Input Layer for External Features} 
We design a 3D CNN module $CNN_{3D}$ to extract temporal representations $\mathbf{f}_T$. This module is constructed using 4 sets of [3D conv - 1D conv]. The first two sets have 3D conv layer with a filter size of 5$\times$5$\times$3 and stride of 3$\times$3$\times$1, and the last two 3D conv layers have a stride of 2$\times$2$\times$1. The final layer merges time channels as one, which results in the output feature of size batch$\times$4$\times$6$\times$512 for frontal view and that of batch$\times$5$\times$5$\times$512 for top-down view. 

The spatial features $\mathbf{f}_S$ are extracted from the stationary environment using a 2D CNN module $CNN_{2D}$. We use 4 sets of [2D conv - 1D conv] where the filter sizes are same as those of $CNN_{3D}$ without depth channel of 3D conv. Note that we use the output of the third set as mid-level semantic context $\Omega_{x^k_\tau}$ to capture the local environment while encoding the motion of the target agent $k$ . 

\vspace{0.5em}
\noindent
\textbf{Input Layer for Node Features} Past motion history $\mathbf{x}^k$ of the target agent $k$ is encoded into high-dimensional feature representations $\mathcal{U}_k$ through MLP with 2 fully connected layers. The resulting features of size batch$\times$512 are added to the corresponding local perception $\Omega_{x^k_\tau}$. We run LSTM and use the last hidden state of size batch$\times512$ to initialize the node feature of the agent $k$, $\mathbf{h}_{(0)}^k$.

For the rest of agents, we follow the similar procedure using MLP and LSTM with the relative motion information. Each individual last hidden state of size batch$\times512$ is used to initialize $\mathbf{h}_{(0)}^j~\forall j\in\{1,...,K\}\setminus\{k\}$.

\vspace{0.5em}
\noindent
\textbf{GNN Layer} The message generation is implemented using three layers of MLP, all with size 256. We use an additional layer of MLP with a same size during the readout phase.  

\subsection{Cross-Modal Embedding}
\noindent
\textbf{Encoder} The ground-truth future motion $\mathbf{y}^k$ of size batch$\times$10$\times$2 is first reshaped to batch$\times$20 and processed through six fully connected layers, each with output size 1024, 6400, 25600, 6400, 1024, and 512. Each layer has a subsequent leaky ReLU layer, and the intermediate features are conditioned by $\mathbf{c}^k$. From the output of the penultimate layer, the first 256 dimension is used as mean and the next 256 dimension is used as standard deviation.

\noindent
\textbf{Decoder} The sampled latent variable is decoded through the 8 fully connected layers. The output size is 1024, 6400, 25600, 102400, 25600, 6400, 1024, and 20, respectively. All layers come together with a ReLU activation function, and each intermediate output is conditioned by $\mathbf{c}^k$.

\end{document}